\documentclass[sigconf,nonacm]{acmart}
\setcopyright{none}               
\renewcommand\footnotetextcopyrightpermission[1]{} 
\usepackage{multirow}
\usepackage{microtype}
\usepackage{pifont}
\usepackage{subcaption}
\AtBeginDocument{%
  }

\begin{document}

\title{Beyond Augmentation: Score-Guided Pathological Prior for EEG-based Depression Detection}

\author{\small Xiaojing Chen$^{1}$, Jingqi Cheng$^{1}$, Xu Zhao$^{1}$, Wan Jiang$^{2}$, Jingjing Wu$^{2}$}
\affiliation{%
  \institution{\small $^{1}$Anhui University \qquad $^{2}$Hefei University of Technology}
  \country{}
}

\renewcommand{\shortauthors}{}
\renewcommand{\shorttitle}{}

\begin{abstract}
Deep learning-based Major Depressive Disorder (MDD) detection using Electroencephalography (EEG) is fundamentally constrained by the small-sample dilemma. Prevailing generative data augmentation methods not only incur heavy computational overhead but also risk introducing synthetic noise, thereby blurring classification boundaries.
To challenge the traditional ``data quantity first'' convention, we propose a novel framework: \textbf{Score-Guided Classification (SGC)}. SGC does not synthesize pseudo-samples; instead, it utilizes an unsupervised generative network architecture to model the structural and statistical anomaly degrees of samples, serving as the core pathological prior. This prior, after robust normalization, is explicitly fused with deep feature representations, thereby precisely guiding the classifier's decision boundary.
Furthermore, to dynamically adapt to varying electrode configurations, we propose a Cross-electrode Spatial Adaptation module, utilizing a spatial mapping mechanism to effectively resolve the hardware heterogeneity of mismatched electrodes in multi-center datasets. Extensive experiments on the Mumtaz2016 and high-density MODMA datasets demonstrate the effectiveness and exceptional generalizability of our method under the challenging zero data augmentation setting and at zero sample synthesis cost.
\end{abstract}



\maketitle

\section{Introduction}
\label{sec:introduction}

\begin{figure}[t]
    \centering
    \includegraphics[width=\linewidth]{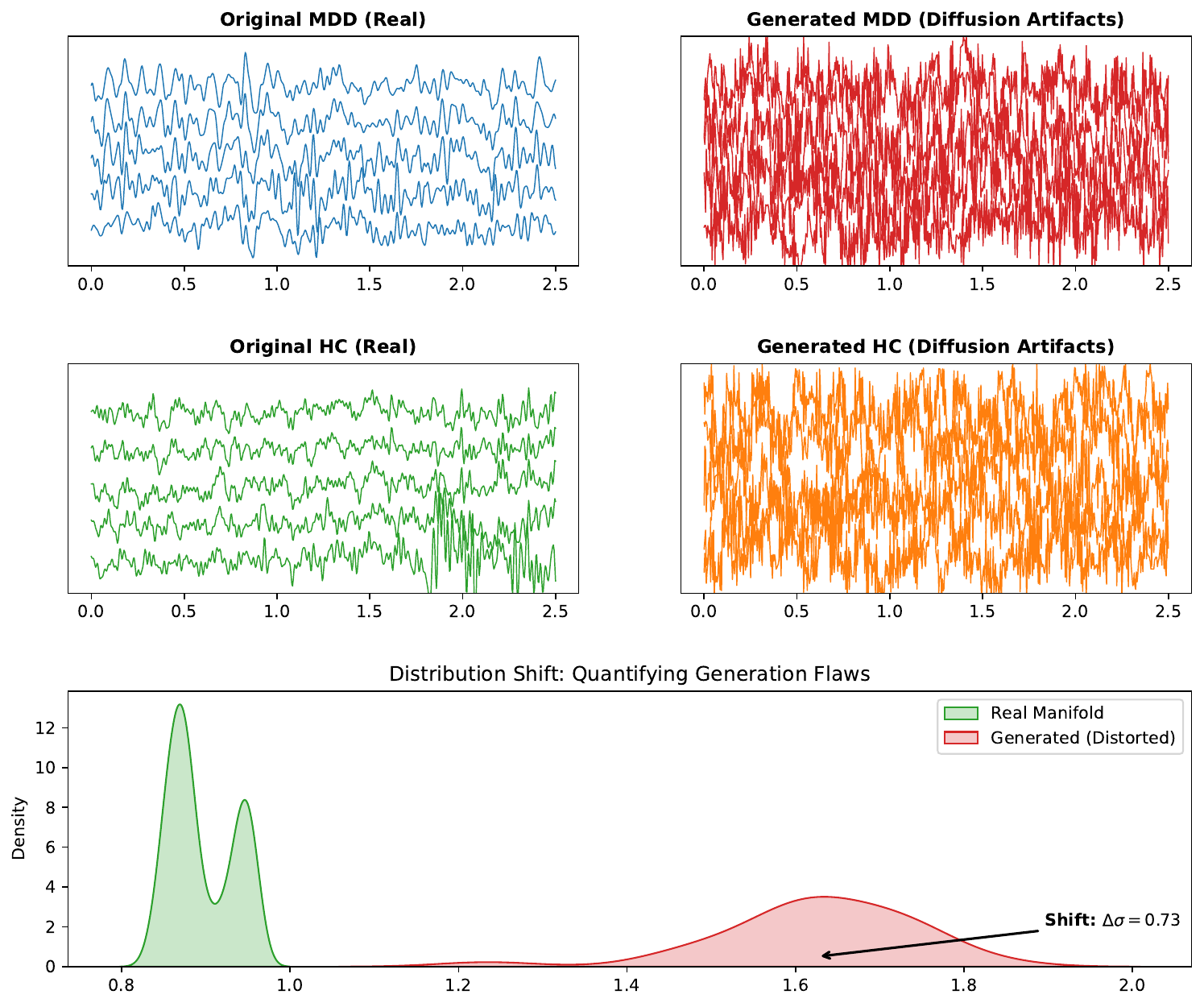} 
    \caption{Generative flaws in standard diffusion-based augmentation. Compared to authentic EEG (top/middle), generated pseudo-signals exhibit severe high-frequency artifacts and morphological distortion. This degradation is quantitatively confirmed (bottom) by a stark distribution shift ($\Delta\sigma = 0.73$) in the KDE plot, demonstrating that synthetic samples fail to preserve the intrinsic statistical manifolds of neural activity.}
    \label{fig:generation_flaw}
\end{figure}

Developing robust computational architectures to decode Electroencephalography (EEG) signals is emerging as a critical frontier for the objective diagnosis of Major Depressive Disorder (MDD) \cite{globaldisability}. This paradigm fundamentally addresses the limitations of traditional psychiatric interviews, which are inherently vulnerable to subjective bias and inter-rater variability. By offering exceptional temporal resolution, EEG provides a direct, non-invasive window into latent cortical dynamics \cite{EEG}. Therefore, the task of EEG-based depression detection has attracted widespread attention in psychiatric informatics and clinical screening.

To process these complex non-stationary signals, contemporary EEG-based MDD detection methodologies generally fall into three representation paradigms. Temporal sequence modeling utilizes 1D Convolutional Neural Networks (1D-CNNs)~\cite{seal2021deprnet} or Recurrent Neural Networks (RNNs) \cite{ay2019automated} to directly extract longitudinal morphological features from raw 1D EEG series. Spatial image transformations map EEG signals into 2D spectrograms or topographical representations to leverage the powerful visual feature extraction capabilities of 2D Convolutional Neural Networks (2D-CNNs) \cite{sharma2023depcap}.Lastly, spatio-temporal and topological modeling employs Graph Neural Networks (GNNs) \cite{luo2023exploring} or Transformers to jointly capture dynamic temporal dependencies and complex cross-electrode functional connectivity \cite{chen2025stge}. By autonomously mining these hierarchical representations, these advanced architectures have significantly elevated the diagnostic performance of automated depression detection.

Despite their impressive performance improvements, these purely end-to-end supervised paradigms exhibit critical vulnerabilities in real-world clinical applications, primarily due to their susceptibility to shortcut learning. EEG signals are inherently plagued by extreme non-stationarity, pronounced inter-subject variability, and exceedingly low signal-to-noise ratios \cite{EEGSignal}. Consequently, mapping these chaotic signals directly to discrete labels forces parameter-heavy networks to degrade into uninterpretable black boxes \cite{blackbox}. Without explicit pathological prior guidance, models tend to overfit subject-specific physiological artifacts rather than isolating universal neurophysiological biomarkers. This overfitting to idiosyncratic noise yields highly brittle decision boundaries, frequently causing a catastrophic degradation in cross-center generalization \cite{lawal2026transfer,wong2026crcc}.

Exacerbating these representational bottlenecks is the severe small-sample dilemma endemic to psychiatric EEG datasets, which typically comprise fewer than 100 subjects \cite{small-sample}. To circumvent this data scarcity, recent literature has increasingly turned to deep generative models, including Graph Neural Networks (GNNs) \cite{luo2023exploring} and Diffusion Models \cite{zhao2024eeg,zhou2023generative}, for brute-force data augmentation \cite{Diffsuion}. However, the direct synthesis of high-dimensional, non-stationary raw EEG signals introduces fatal methodological flaws. As compellingly demonstrated in Fig.~\ref{fig:generation_flaw}, the directly synthesized pseudo-signals are plagued by dense, high-frequency generative artifacts and severe morphological distortions \cite{zhang2026stats}. Crucially, these hallucinated samples exhibit a stark distribution shift ($\Delta\sigma = 0.73$) away from the authentic data manifold. Consequently, this noisy augmentation paradigm actively injects spurious correlations, misleading downstream classifiers into memorizing synthetic noise distributions rather than clinically meaningful biomarkers \cite{choi2026fail}.

To fundamentally circumvent the representational bottlenecks of discriminative black boxes and the catastrophic noise injection endemic to generative augmentation, we propose a novel framework that steps beyond augmentation: \textbf{S}core-\textbf{G}uided \textbf{C}lassification (\textbf{SGC}). Distinct from existing paradigms that blindly hallucinate raw EEG, SGC redefines generative utility by modeling a normative electrophysiological baseline. Specifically, we design an unsupervised dual-stream generative pipeline trained exclusively on healthy control (HC) cohorts to anchor the intrinsic manifold of normal brain dynamics. To comprehensively capture this normative baseline, the dual-stream architecture is meticulously constructed: one stream explicitly models the discrete structural morphology of healthy signals, while the complementary stream captures their continuous probabilistic distribution. For any input, SGC quantifies the structural reconstruction error and probabilistic distribution shift against this frozen template, distilling a fine-grained anomaly score \cite{rutherford2022normative}. Serving as a deterministic, noise-free pathological prior, this continuous score is explicitly fused into the downstream deep classifier to dynamically calibrate the decision boundary, effectively isolating genuine MDD biomarkers from complex background variance.

Beyond representational challenges, the real-world clinical deployment of EEG-based diagnostics is frequently bottlenecked by hardware-level data heterogeneity, such as varying electrode configurations across clinical centers, ranging from 19 to 128 channels. To equip the SGC architecture with hardware-agnostic generalizability, we introduce a continuous spatial topology mapping strategy. By geometrically projecting arbitrary high-density raw signals onto a unified, standardized electrode layout, this mechanism maximally preserves the global biophysical topography of the brain. Crucially, it eradicates the necessity for architecture-level structural retraining, thereby unlocking seamless zero-shot transfer and robust cross-dataset adaptation across highly heterogeneous clinical environments.

In summary, our contributions are:

\ding{113} We propose \textbf{Score-Guided Classification (SGC)}, a novel augmentation-free framework that models a normative EEG baseline from healthy controls via an unsupervised dual-stream architecture, and exploits the resulting pathological prior to enhance downstream depression detection by  reshaping feature manifolds and broadening decision margins.

\ding{113} We introduce a spatial topology mapping strategy that projects mismatched electrodes onto a unified layout. This resolves cross-cohort hardware heterogeneity and unlocks seamless zero-shot transfer without architectural retraining.

\ding{113} Extensive evaluations on Mumtaz2016 and MODMA yield state-of-the-art performance (95.19\% accuracy), empirically validating that extracting explicit pathological priors is fundamentally more robust than standard generative augmentation.
\begin{figure*}[t]
  \centering
  \includegraphics[width=0.90\textwidth, trim=2.8cm 3cm 1.8cm 3.2cm, clip]{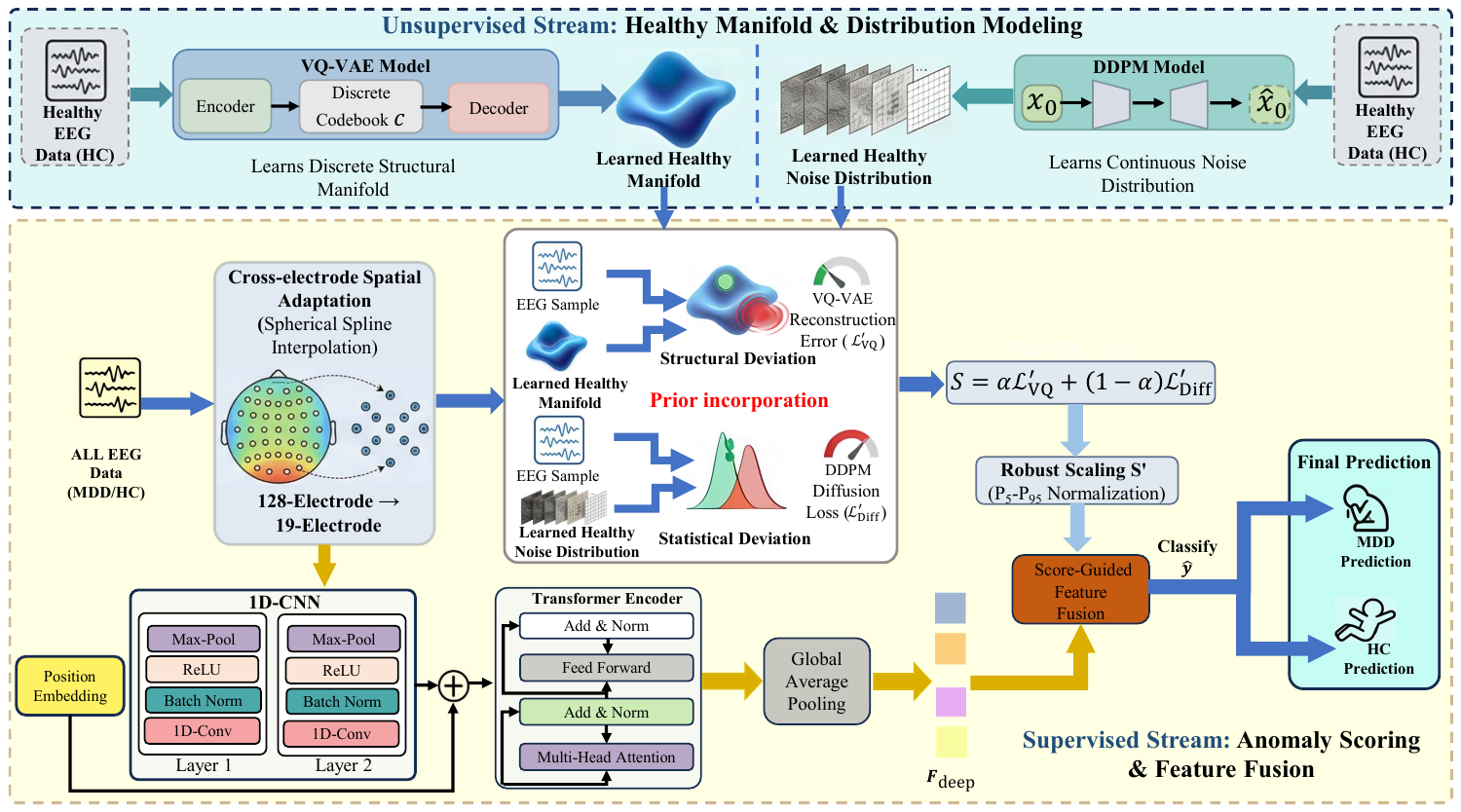} 
  
  \vspace{-0.3cm} 
  
  \caption{Overall architecture of the Score-Guided Classification (SGC) framework. The \textit{unsupervised stream} (top) exclusively models the normative healthy baseline using VQ-VAE and DDPM. The \textit{supervised stream} (bottom) utilizes this frozen baseline to extract a pathological anomaly score ($S'$) for incoming samples, while in parallel capturing deep spatio-temporal features ($F_{\text{deep}}$). Late-stage feature fusion integrates $S'$ to precisely guide the final decision boundary.}
  \label{fig:framework}

  \vspace{-0.2cm} 
\end{figure*}

\section{Related Work}
\label{sec:related_work}

\subsection{EEG-based Depression Detection}
The landscape of automatic Major Depressive Disorder (MDD) detection has experienced a significant paradigm shift from traditional handcrafted spectral feature engineering to data-driven deep learning architectures. Recent literature has extensively explored various deep frameworks, including 1D Convolutional Neural Networks (1D-CNNs), Graph Neural Networks (GNNs), and Transformer-based models, to capture complex spatio-temporal dynamics and cross-channel topological dependencies within EEG signals \cite{mastgcn_mdd, cnn_transformer_mdd_2025,singh2024slitranet,wang2024gctnet,hou2025lightweight,11369854}. Despite their impressive representational capabilities, these purely supervised, parameter-heavy paradigms are notoriously susceptible to severe overfitting, particularly when deployed in the small-sample regimes endemic to clinical psychiatric datasets \cite{eeg_review_overfitting_2026,liu2022machine,vaniya2026simplifying,olbrich2026deep}. Consequently, they frequently exhibit substantial performance degradation in subject-independent generalization scenarios \cite{shen2025wdanet,wang2024major,li2024distillation,kim2025domain}. This inherent limitation underscores the critical necessity for novel representation learning strategies that transcend the boundaries of simplistic end-to-end supervised classification.

\subsection{Data Augmentation for EEG Analysis}To mitigate the pervasive data scarcity in EEG-based assessments, extensive research has been dedicated to data augmentation techniques, which broadly bifurcate into heuristic and generative approaches. Traditional heuristic methods, such as random cropping, temporal masking, or signal flipping, attempt to expand the training manifold but frequently violate the intricate biophysical integrity of EEG recordings, inadvertently inducing semantic distortions and phase-coupling disruptions \cite{rommel2022data,lashgari2020data,liao2025eeg,lee2026rl}. To circumvent these handcrafted limitations, recent advancements have increasingly leveraged deep generative architectures, notably Generative Adversarial Networks (GANs)~\cite{hartmann2018eeg} and Denoising Diffusion Probabilistic Models (DDPMs)~\cite{zhao2024eeg,zhou2023generative}, to synthesize high-fidelity pseudo-samples. However, deploying these generative frameworks in ultra-low-data scenarios introduces severe methodological bottlenecks \cite{11328798}. As empirically demonstrated in our preliminary analysis (see Fig.~\ref{fig:generation_flaw}), diffusion-based synthesis on limited EEG cohorts incurs pronounced high-frequency noise amplification and a stark distribution shift ($\Delta\sigma$) away from the authentic data manifold \cite{he2021data,lin2024diffusion}. Rather than sharpening decision boundaries, such noisy data expansion irreparably blurs them. In contrast to these augmentation-centric paradigms, our SGC framework fundamentally shifts the focus toward unsupervised feature excavation, completely bypassing the risks associated with hallucinated raw signal generation.

\subsection{Anomaly Detection as Pathological Priors}
Conventional Anomaly Detection (AD) frameworks often reduce pathological deviations to rigid thresholds for outlier rejection \cite{fernando2021deep,an2015variational,baur2018deep}, underutilizing generative representations. Their reliance on holistic reconstruction errors or shallow statistical distances is easily confounded by EEG's extreme non-stationarity and inter-subject variability. Departing from this rigid paradigm, SGC redefines anomaly scores as continuous pathological priors. By synergizing structural quantization \cite{van2017neural} and probabilistic diffusion \cite{ho2020denoising}, we construct a unified prior space isolating anomalies across complementary dimensions. Specifically, quantization identifies morphological distortions in temporal microstates, while diffusion captures subtle statistical drifts in background dynamics. This calibration dynamically reshapes the downstream decision boundary, transforming rudimentary outlier metrics into highly discriminative features for robust MDD diagnostics.

\section{Methodology}
\label{sec:methodology}

\subsection{Problem Formulation and Dual-Stream Architecture}
\label{subsec:framework_overview}

We formulate EEG-based depression detection as a prior-guided time-series classification task. Given a dataset $\mathcal{D} = \{(\mathbf{x}_i, y_i)\}_{i=1}^N$, where $\mathbf{x}_i \in \mathbb{R}^{C \times T}$ denotes an EEG segment comprising $C$ channels and $T$ time steps, and $y_i \in \{0, 1\}$ represents the ground-truth label (0 for HC, 1 for MDD), our Score-Guided Classification (SGC) framework aims to learn an optimized mapping function $\mathcal{F}: \mathbf{x}_i \rightarrow \hat{y}_i$. Departing from conventional end-to-end paradigms, SGC dynamically calibrates the classifier's decision boundary by explicitly injecting a robust pathological anomaly score $S' \in \mathbb{R}$, which is derived independently via unsupervised generative modeling.

As illustrated in Fig.~\ref{fig:framework}, the SGC framework is operationalized through a synergistic dual-stream architecture:

\ding{113} {Unsupervised Stream (Healthy Manifold \& Distribution Modeling):} Trained exclusively on normative HC data, this offline generative branch establishes a standard structural manifold and a baseline noise distribution to completely anchor the intrinsic patterns of healthy neural dynamics.

\ding{113} {Supervised Stream (Anomaly Scoring \& Feature Fusion):} Operating on incoming samples, this discriminative pipeline first standardizes hardware heterogeneity via cross-electrode spatial adaptation. It then leverages the frozen unsupervised models to quantify the pathological deviation of the input (yielding the anomaly score $S'$), and explicitly fuses this score with deeply extracted spatiotemporal features to guide the final classification.

The fundamental insight of this architecture lies in its late-stage feature fusion. By leveraging unsupervised generative models to anchor the intrinsic normative patterns, SGC constructs a deterministic, noise-free pathological reference. This continuous prior explicitly steers the final Multi-Layer Perceptron (MLP) mapping, shielding the network from spurious correlations and ensuring highly robust MDD predictions. The subsequent subsections detail the spatial adaptation strategy for hardware heterogeneity, the specific designs of these two streams, and the overall training optimization.
\subsection{Cross-electrode Spatial Adaptation}
\label{subsec:spatial_adaptation}

Cross-cohort EEG analysis is frequently hindered by heterogeneous electrode configurations, such as high-density 128-electrode arrays versus standard 19-electrode clinical systems.To ensure the broad generalizability of SGC without necessitating architectural modifications or costly structural retraining, we formulate a spatial topology mapping strategy.

Specifically, we project the arbitrary high-density signals $\mathbf{X}_{\text{orig}} \in \mathbb{R}^{C_{\text{orig}} \times T}$ onto a standardized 19-electrode spatial template (following the International 10-20 system \cite{klem1999ten}) via \textbf{Spherical Spline Interpolation} \cite{perrin1989spherical}. By mathematically mapping each electrode to a 3D coordinate $\mathbf{p}$ on a normalized spherical manifold representing the human scalp, the interpolated voltage at any target electrode $\mathbf{p}_{target}$ is estimated using the original source electrodes $\mathbf{p}_j$:
\begin{equation}
    f(\mathbf{p}_{\text{target}}) = \sum_{j=1}^{C_{\text{orig}}} c_j g(\|\mathbf{p}_{\text{target}} - \mathbf{p}_j\|)
\label{eq:spline_interpolation}
\end{equation}
\noindent where $g(\cdot)$ denotes the spherical spline Green's function, and $c_j$ represents the spatial interpolation coefficients. This geometric projection strictly enforces a consistent spatial dimensionality ($C=19$) for the downstream dual-stream network, thereby unlocking seamless cross-dataset zero-shot inference capabilities.

\subsection{Unsupervised Healthy Manifold and Distribution Modeling}
\label{subsec:unsupervised_modeling}

We design unsupervised healthy manifold and distribution modeling branch to extract pathological prior from healthy samples.
To this end, it excavates normative EEG characteristics via a one-class generative modeling paradigm. Modeling the authentic healthy EEG manifold poses significant challenges due to the intrinsic duality of the signals, which comprise discrete structural microstates, such as quasi-stable physiological waveforms, and continuous stochastic background dynamics. Monolithic generative models, including vanilla VAEs \cite{kingma2013auto} and GANs \cite{goodfellow2014generative}, typically fail to capture both facets simultaneously, frequently suffering from blurry reconstructions or mode collapse. To overcome this representational bottleneck, we establish a robust \textbf{Healthy Manifold} \cite{marquand2016understanding} by training exclusively on Healthy Control (HC) data through a meticulously decoupled dual-stream pipeline.

To handle these issues, first, we employ a Vector-Quantized Variational Autoencoder (VQ-VAE) \cite{van2017neural,yang2025thd} as a {Manifold Learner}. Its discrete codebook mechanism is fundamentally suited to quantize and memorize the structural integrity of normative EEG patterns. Complementarily, a Denoising Diffusion Probabilistic Model (DDPM) \cite{ho2020denoising,pinaya2022fast} serves as a \textit{Noise Predictor}. Its continuous Markovian diffusion process is uniquely optimal for modeling the complex, high-dimensional probability distribution of the stochastic background. By establishing this dual baseline, the unsupervised stream perfectly encapsulates the normative electrophysiological state, laying the definitive foundation for quantifying pathological deviations in the subsequent supervised classification phase.

\subsubsection{Structure Reconstruction }
\label{subsubsec:vq_vae}

To effectively encode the complex, non-stationary temporal dynamics of EEG signals, we employ a 1D-CNN-based VQ-VAE architecture. The encoder $E(\cdot)$ projects the input signal $\mathbf{x} \in \mathbb{R}^{C \times T}$ into a continuous latent space, yielding the representation $\mathbf{z}_e$. The quantizer $Q(\cdot)$ then discretizes $\mathbf{z}_e$ into $\mathbf{z}_q$ via nearest-neighbor mapping within a learnable codebook $\mathcal{C} = \{\mathbf{e}_k\}_{k=1}^{K}$ \cite{van2017neural}. Crucially, this discrete codebook ($K=512$, with a dimensionality of $64$) is optimized exclusively on HC data. This constraint forces the network to map input signals into a finite set of \textbf{healthy basis atoms}, effectively constructing a robust structural ``dictionary'' of normative EEG patterns.

The decoder $D(\cdot)$ subsequently reconstructs the signal from the quantized vector $\mathbf{z}_q$. To optimize this healthy manifold, the network is trained using the VQ-VAE objective $\mathcal{L}_{\text{VQ}}$, which combines the Mean Squared Error (MSE) for reconstruction fidelity with a codebook commitment loss:
\begin{equation}
    \mathcal{L}_{\text{VQ}} = \|\mathbf{x} - D(\mathbf{z}_q)\|_2^2 + \beta \|\text{sg}[\mathbf{z}_e] - \mathbf{z}_q\|_2^2
    \label{eq:vq_loss}
\end{equation}
\noindent where $\text{sg}[\cdot]$ denotes the stop-gradient operator, and the commitment cost $\beta$ is empirically set to $0.25$ \cite{van2017neural}. By strictly bounding the optimization landscape to normative HC data, $\mathcal{L}_{\text{VQ}}$ is uniquely positioned to act as a rigorous metric: any subsequent out-of-distribution sample will inevitably incur a high reconstruction error when forced through this healthy dictionary.

\subsubsection{Distributional Probability Modeling}
\label{subsubsec:ddpm}

To complement the VQ-VAE's discrete representations, we independently train a Denoising Diffusion Probabilistic Model (DDPM) \cite{ho2020denoising} to map the continuous probabilistic landscape of healthy EEG signals. The forward phase gradually corrupts a normative signal $\mathbf{x}_0$ by injecting Gaussian noise across $T=500$ steps. At timestep $t$, the perturbed state is:
\begin{equation}
    q(\mathbf{x}_t \mid \mathbf{x}_0) = \mathcal{N}\left(\mathbf{x}_t; \sqrt{\bar{\alpha}_t} \mathbf{x}_0, (1 - \bar{\alpha}_t)\mathbf{I} \right),
    \label{eq:diffusion_forward}
\end{equation}
\noindent where $\{\bar{\alpha}_t\}_{t=1}^{T}$ dictates the variance schedule.

Conversely, the reverse generative process trains a parameterized network $\epsilon_\theta$ to estimate the injected noise $\epsilon$. To exclusively encapsulate the normative distribution, the network is optimized on HC data using the diffusion objective $\mathcal{L}_{\text{Diff}}$ \cite{ho2020denoising}:
\begin{equation}
    \mathcal{L}_{\text{Diff}} = \mathbb{E}_{t, \mathbf{x}_0, \epsilon \sim \mathcal{N}(0, \mathbf{I})}\left[ \| \epsilon - \epsilon_\theta(\mathbf{x}_t, t) \|_2^2 \right].
    \label{eq:diff_loss}
\end{equation}
By minimizing this loss, the DDPM learns the inherent stochastic baseline of healthy neurodynamics. Consequently, $\mathcal{L}_{\text{Diff}}$ establishes a rigorous foundation to measure implicit statistical shifts in subsequent out-of-distribution samples, perfectly complementing the VQ-VAE.

\subsection{Supervised Stream: Anomaly Scoring and Feature Fusion}
\label{subsec:supervised_fusion}

To fully exploit the complex spatiotemporal dynamics of EEG signals and explicitly leverage the learned normative baseline, we design a comprehensive supervised classification stream. As depicted in the lower branch of our overall architecture, this phase shifts from offline generative modeling to online inference and classification. It sequentially performs three synergistic operations: anomaly score extraction via the frozen generative models, deep feature representation using a hybrid local-global backbone, and score-guided feature fusion.

Crucially, rather than relying solely on raw morphological representations ($F_{\text{deep}}$), we introduce a feature-level fusion mechanism that explicitly injects the normalized anomaly score ($S'$) as a deterministic soft prior. By concatenating $S'$ with $F_{\text{deep}}$, we construct an augmented, prior-guided representation ($F_{\text{final}}$) to calibrate the final Multi-Layer Perceptron (MLP) classifier. This dual-awareness mechanism explicitly forces the network to focus on discriminative pathological patterns rather than idiosyncratic background noise.

\subsubsection{Pathological Prior Incorporation (Anomaly Scoring)}
\label{subsubsec:score_extraction}

Specifically, SGC leverages the normative baselines established in the unsupervised stream (Section~\ref{subsec:unsupervised_modeling}). For an incoming EEG segment $\mathbf{x}$, we employ the frozen pre-trained VQ-VAE and DDPM architectures to act as fixed reference templates. By projecting $\mathbf{x}$ through these frozen models, we compute the inference-stage structural error $\mathcal{L}'_{\text{VQ}}$ (via Eq.~\ref{eq:vq_loss}) and the statistical deviation $\mathcal{L}'_{\text{Diff}}$ (via Eq.~\ref{eq:diff_loss}), respectively. These metrics quantify the input's pathological deviation from the learned healthy manifold and noise distribution. To synthesize these explicit structural and implicit statistical deviations, we compute a composite anomaly score $S$:
\begin{equation}
    S = \alpha \mathcal{L}'_{\text{VQ}} + (1 - \alpha) \mathcal{L}'_{\text{Diff}},
    \label{eq:anomaly_score_raw}
\end{equation}
\noindent where $\alpha=0.7$ determined via a held-out pilot split. This higher weight favors the deterministic VQ-VAE error over the stochastic DDPM loss, ensuring the stability of the final anomaly score.

Because raw scores are highly susceptible to inter-subject variability and extreme outliers, we apply robust normalization to project $S$ into a bounded final prior $S'$ \cite{truong2025data,el2020automatic}:
\begin{equation}
    S' = \operatorname{clip}\left(\frac{S - P_5}{P_{95} - P_5}, 0, 1\right),
    \label{eq:robust_scaling}
\end{equation}
\noindent where $P_5$ and $P_{95}$ are training percentiles explicitly filtering the unpredictable 5\% physiological noise tails.

\begin{figure}[h] 
    \centering
    \includegraphics[width=0.95\linewidth]{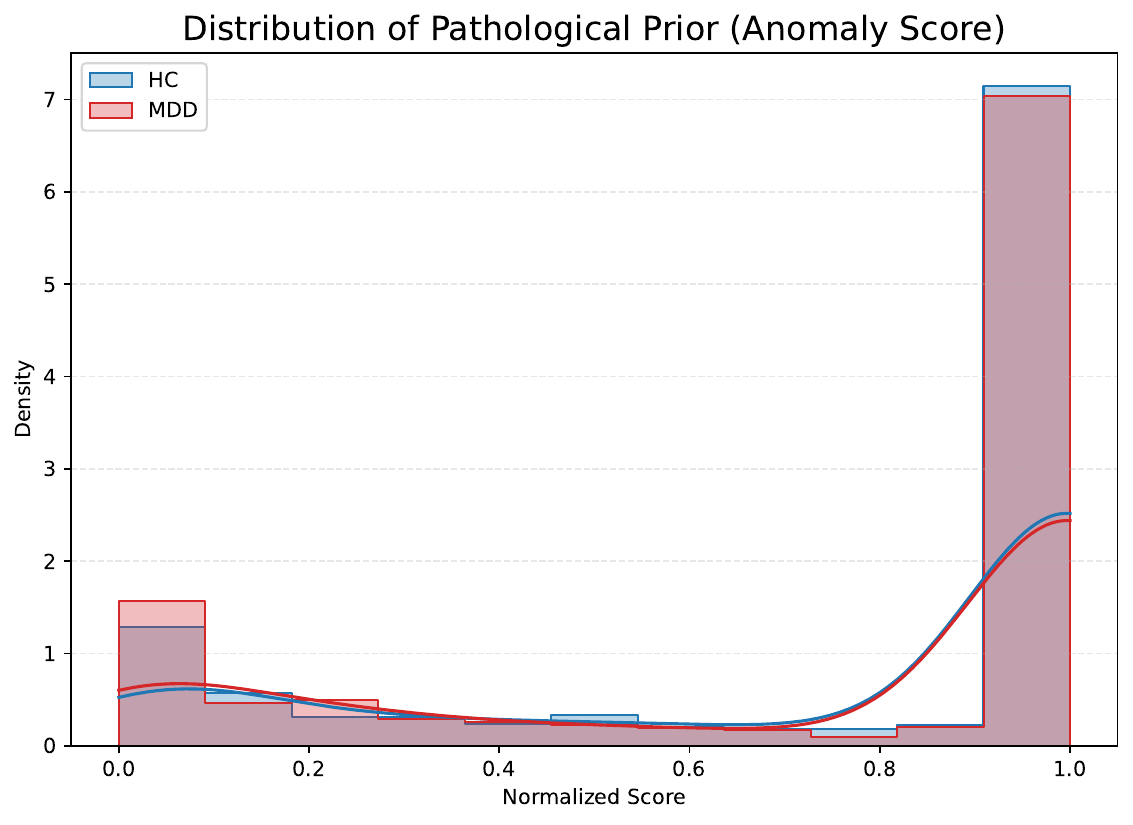} 
    \caption{Probability density distribution of the normalized anomaly scores ($S'$). The inevitable boundary overlap motivates the integration of this score as a soft prior rather than a brittle hard threshold.}
    \label{fig:score_dist}
\end{figure}

As shown in Fig.~\ref{fig:score_dist}, while the MDD cohort exhibits substantially higher scores, the intrinsic non-stationarity of EEG induces inevitable boundary overlaps. Consequently, rather than enforcing a rigid, error-prone diagnostic threshold, we seamlessly embed $S'$ into the network as a continuous \textbf{soft prior feature} \cite{choudhury2025improving,zhang2023soft}. This mapping acts as an informative anchor, explicitly signaling the magnitude of pathological deviation to the downstream classifier.

\subsubsection{Hybrid Feature Extractor }
\label{subsubsec:hybrid_extractor}

The backbone extracts complex EEG characteristics through a hierarchical, two-stage architecture. Initially, a 1D-CNN frontend is deployed to capture local temporal morphological features, such as high-frequency oscillations and micro-state peaks. The raw input segment is iteratively updated via 1D convolution, Batch Normalization (BN) to mitigate internal covariate shifts \cite{ioffe2015batch}, ReLU activation, and Max Pooling to progressively broaden the temporal receptive field and filter redundant noise.

Subsequently, the downsampled CNN feature maps are sequence-projected and fed into a Transformer Encoder. To comprehensively model the long-range global dependencies across the entire temporal dimension, we employ the standard Multi-Head Self-Attention (MHSA) mechanism \cite{vaswani2017attention}. Finally, a Global Average Pooling (GAP) layer aggregates the output token sequence into a compact, discriminative deep spatiotemporal feature vector, denoted as $\mathbf{F}_{\text{deep}}$.

\subsubsection{Score-Guided Feature Fusion}
\label{subsubsec:feature_fusion}

Unlike conventional classifiers relying solely on raw signals, SGC incorporates the normalized anomaly score $S'$ as an explicit \textit{pathological prior}. Before the final Multi-Layer Perceptron (MLP) head, $S'$ is directly concatenated ($\oplus$) with the deep spatiotemporal feature vector $F_{\text{deep}}$:
\begin{equation}
    F_{\text{final}} = F_{\text{deep}} \oplus S'
    \label{eq:feature_fusion}
\end{equation}
This deterministic fusion endows the classifier with dual-awareness, simultaneously perceiving morphological representations and statistical deviations. By treating the scalar prior as an influential trigger alongside high-dimensional features, it sharply refines the decision boundary, effectively classifying challenging samples where ambiguous morphology is clarified by pathological scores.

\begin{table*}[t]
\centering
\caption{Performance comparison with State-of-the-Art (SOTA) methods on Mumtaz2016 (10-fold cross-validation). Note: All baselines were re-implemented and evaluated under our unified zero-shot pipeline, resulting in slight numerical deviations from original reports.}
\label{tab:sota_comparison}
\resizebox{\textwidth}{!}{
\begin{tabular}{l|ccccc|ccccc}
\toprule
\multirow{2}{*}{\textbf{Methods}} & \multicolumn{5}{c|}{\textbf{Window-level Metrics (\%)}} & \multicolumn{5}{c}{\textbf{Subject-level Metrics (\%)}} \\
\cmidrule(lr){2-6} \cmidrule(lr){7-11}
 & \textbf{Acc} & \textbf{Prec} & \textbf{Recall} & \textbf{F1} & \textbf{AUC} & \textbf{Acc} & \textbf{Prec} & \textbf{Recall} & \textbf{F1} & \textbf{AUC} \\
\midrule
LR [15]               & 81.63 & 81.68 & 79.07 & 80.02 & 88.28 & 83.33 & 84.67 & 78.08 & 80.58 & 90.54 \\
SVM [15]              & 81.13 & 82.13 & 76.97 & 79.20 & 87.31 & 81.52 & 82.67 & 76.08 & 78.58 & 88.91 \\
XGBoost [15]          & 83.25 & 83.54 & 82.41 & 82.49 & 88.73 & 89.17 & 93.71 & 86.33 & 89.03 & 92.19 \\
1D-CNN-LSTM [2]       & 90.53 & 90.17 & 91.58 & 90.52 & 93.13 & 93.26 & 93.25 & 95.83 & 94.01 & 95.13 \\
1D-CNN-Transformer [41] & 85.56 & 84.90 & 87.97 & 84.98 & 92.39 & 86.52 & 86.02 & 89.63 & 86.21 & 93.96 \\
CWT-1D-CNN [44]       & 91.54 & 91.95 & 92.30 & 91.53 & 92.57 & 93.18 & 92.80 & 95.00 & 93.31 & 95.42 \\
CWT-2D-CNN [44]       & 90.50 & 89.17 & 93.63 & 90.87 & 94.49 & 94.09 & 94.46 & 95.50 & 94.54 & 96.78 \\
EEGNet [25]           & 89.75 & 90.48 & 91.32 & 90.45 & 94.29 & 91.67 & 91.53 & 93.75 & 92.04 & 95.66 \\
InceptionNet [52]     & 90.54 & 88.26 & 93.51 & 91.38 & 93.15 & 94.92 & 90.46 & 95.27 & 94.62 & 94.69 \\
TSception [8]         & 90.23 & 89.08 & 91.74 & 89.90 & 96.10 & 91.67 & 92.14 & 92.08 & 91.38 & 96.70 \\
GRU-Conv [64]         & 90.95 & 90.45 & 90.92 & 90.32 & 96.19 & 92.42 & 93.25 & 90.83 & 90.58 & 97.45 \\
DeprNet [47]          & 90.00 & 90.66 & 91.02 & 90.09 & 97.59 & 91.59 & 92.10 & 93.75 & 92.02 & 98.22 \\
GC-GRU [34]           & 89.74 & 90.09 & 87.90 & 88.78 & 96.01 & 93.26 & 93.46 & 91.00 & 91.79 & 97.90 \\
DiffMDD [61]          & 93.85 & 92.52 & 94.86 & 93.65 & 95.31 & 96.55 & \textbf{96.21} & 94.33 & 93.84 & 98.62 \\
\midrule
\textbf{SGC (Ours)}   & \textbf{95.28} & \textbf{93.45} & \textbf{97.18} & \textbf{95.18} & \textbf{98.83} & \textbf{96.67} & 95.79 & \textbf{97.50} & \textbf{96.30} & \textbf{100.00} \\
\bottomrule
\end{tabular}
}
\end{table*}

\subsection{Training Strategy Optimization}
\label{subsec:training_strategy}

To counteract the severe overfitting and convergence instability endemic to small-sample EEG regimes, we devise a robust optimization protocol.

\subsubsection{Optimizer and Regularization}
\label{subsubsec:optimizer_regularization}

We institute a dual-pronged regularization scheme. First, we discard standard step-wise decay in favor of a Cosine Annealing learning rate scheduler \cite{loshchilov2016sgdr}. By periodically injecting energy into the optimization trajectory, this approach effectively propels the model out of sharp, spurious local minima, guiding the weights toward flatter regions of the loss landscape that inherently possess superior generalization bounds.

Concurrently, to penalize the classifier's tendency toward over-confidence on limited training manifolds, we introduce Label Smoothing \cite{szegedy2016rethinking,muller2019does}. This technique relaxes the rigid one-hot ground-truth labels $y_i$ into continuous soft targets $\tilde{y}_i$:
\begin{equation}
    \tilde{y}_i = (1 - \delta) \cdot y_i + \frac{\delta}{K},
    \label{eq:label_smoothing}
\end{equation}
\noindent where the smoothing margin $\delta$ is strictly set to $0.1$ for our binary classification setting ($K=2$).

\subsubsection{Supervised Classification Loss}
\label{subsubsec:classification_loss}

The end-to-end supervised stream is optimized via a smoothed binary cross-entropy objective. For a mini-batch of $N$ predictions $\hat{y}_i$ and their corresponding softened targets $\tilde{y}_i$ (Eq.~\ref{eq:label_smoothing}), the classification loss $\mathcal{L}_{\text{cls}}$ is formalized as:
\begin{equation}
    \mathcal{L}_{\text{cls}} = -\frac{1}{N} \sum_{i=1}^{N} \left[ \tilde{y}_i \log(\hat{y}_i) + (1 - \tilde{y}_i) \log(1 - \hat{y}_i) \right].
    \label{eq:cls_loss}
\end{equation}
Synergizing this objective with the aforementioned regularization strategies strictly bounds the hypothesis space, ensuring stable convergence while effectively neutralizing the risk of memorizing dataset-specific noise.

\section{Experiments and Results}
\label{sec:experiments}

\subsection{Experimental Setup}
\label{subsec:experimental_setup}

\subsubsection{Datasets and Preprocessing}
\label{subsubsec:datasets}

\noindent\textbf{Mumtaz2016 \cite{mumtaz2017electroencephalogram}:} This benchmark comprises resting-state, 19-electrode EEG recordings from 34 MDD patients and 30 HCs. Preprocessing includes a 0.5-50.0 Hz band-pass filter, 256 Hz resampling, and 5-second sliding window segmentation (50\% overlap), yielding 13,953 segments. Under our subject-independent 10-fold cross-validation scheme, each fold partitions the data into $\sim$12,557 training and $\sim$1,396 testing segments. To ensure zero data leakage, the unsupervised stream is independently retrained for each fold using only the healthy segments within the current training set ($\sim$6,038 HC segments), while the supervised stream utilizes the entire $\sim$12,557-segment training set.

\noindent\textbf{MODMA \cite{cai2022multi}:} Serving as a rigorous cross-domain testbed, this dataset contains 128-electrode high-density EEG from 24 MDD and 29 HCs. Following artifact removal, we apply our Spatial Topology Mapping to geometrically project the 128 electrodes onto the standard 19-electrode layout. Identical segmentation generates 6,355 samples, which are strictly reserved as an unseen zero-shot test set for final evaluation.

\subsubsection{Implementation Details}
\label{subsubsec:implementation_metrics}

Models are implemented in PyTorch on an NVIDIA RTX 6000 GPU, optimized via Adam and a Cosine Annealing scheduler. The unsupervised stream configures the VQ-VAE codebook ($K=512$, dimension $64$) and DDPM ($T=500$, linear noise $\beta \in [10^{-4}, 0.02]$). The supervised stream uses an initial learning rate of $5 \times 10^{-4}$, weight decay of $1 \times 10^{-4}$, and $0.5$ dropout. Label smoothing ($\delta=0.1$) and early stopping (patience 15) are employed to mitigate overfitting.

 \textbf{Segment-level metrics} 
 that evaluate individual EEG windows include Accuracy (Acc), Precision, F1-Score, AUC, and Recall (Rec). \textbf{Subject-level Accuracy (Sub-Acc)} determines the final patient-level diagnosis via majority voting across an individual's segments.


\subsection{Comparison with State-of-the-Art}
\label{subsec:sota_comparison}

To rigorously evaluate SGC on the Mumtaz2016 dataset, we benchmark against 14 established methods spanning four paradigms (detailed citations are provided in Table~\ref{tab:sota_comparison}). These include traditional machine learning methods such as LR, SVM, and XGBoost; deep discriminative models like EEGNet, InceptionNet, and DeprNet; spatio-temporal architectures including 1D-CNN-LSTM, 1D-CNN-Transformer, CWT-CNNs, TSception, GRU-Conv, and GC-GRU; as well as generative augmentation approaches represented by DiffMDD. For fairness, all deep learning baselines are re-implemented and evaluated under our identical 10-fold cross-validation and zero-shot pipelines.

As detailed in Table~\ref{tab:sota_comparison}, SGC consistently dominates across all metrics. Compared to traditional 1D-CNNs (DeprNet: 89.84\%) and spatio-temporal networks (GC-GRU: 90.15\%), SGC achieves 95.19\% accuracy without generating a single new sample. Crucially, it outperforms the SOTA generative augmentation model DiffMDD (93.85\%). This compellingly demonstrates that explicitly extracting high-quality pathological priors is fundamentally more effective than incurring the computational costs and noise risks of pseudo-sample generation.

\subsection{Cross-Dataset Generalization: Robustness to Electrode Heterogeneity}
\label{subsec:generalization}

To rigorously evaluate SGC's zero-shot clinical transferability, we conduct external validation on the MODMA dataset without any domain-specific fine-tuning. This presents a severe \textit{domain shift} due to extreme hardware disparities (128-electrode to 19-electrode interpolation) and distinct patient demographics. The results are detailed in Table~\ref{tab:generalization_modma}.

\begin{table}[h]
  \centering
  \caption{Cross-dataset zero-shot generalization (MODMA). Traditional SOTA models and generative augmentation suffer severe domain shift, while SGC remains robust.}
  \label{tab:generalization_modma}
  
  \small 
  \setlength{\tabcolsep}{2pt} 
  
  \renewcommand{\arraystretch}{1.2}
  
  \resizebox{\linewidth}{!}{
  \begin{tabular}{l c c c c c}
    \toprule
    \textbf{Method} & \textbf{Acc (\%)} & \textbf{Pre (\%)} & \textbf{Rec (\%)} & \textbf{F1 (\%)} & \textbf{Sub-Acc (\%)} \\
    \midrule
    1D-CNN-LSTM \cite{ay2019automated}  & 57.20 & 54.35 & 58.12 & 56.18 & 58.50 \\
    CWT-1D-CNN \cite{raghavendra2023automated} & 59.80 & 57.66 & 57.34 & 58.47 & 60.75 \\
    XGBoost   \cite{hassan2024impact}       & 58.45 & 55.12 & 60.30 & 57.60 & 59.80 \\
    TSception \cite{ding2020tsception}    & 60.12 & 56.45 & 59.85 & 58.10 & 61.25 \\
    DeprNet \cite{seal2021deprnet}      & 56.80 & 52.30 & 58.42 & 55.18 & 58.33 \\
    GC-GRU \cite{luo2023exploring}       & 55.40 & 51.20 & 56.75 & 53.82 & 56.00 \\
    DiffMDD \cite{wang2024diffmdd}       & 68.34 & 64.32 & 58.60 & 58.94 & 69.36 \\
    \midrule
    \textbf{SGC (Ours)}      & \textbf{78.50} & \textbf{70.82} & \textbf{75.34} & \textbf{69.96} & \textbf{79.94} \\
    \bottomrule
  \end{tabular}
  } 
\end{table}

\noindent\textbf{1) Cross-Domain Brittleness of Existing Paradigms.} Purely data-driven methods degrade catastrophically in zero-shot transfer. Notably, GC-GRU plummets to 55.40\% accuracy, as cross-hardware spatial interpolation disrupts source-domain functional topologies. Furthermore, even the state-of-the-art generative augmentation model, DiffMDD (68.34\%), exhibits significant brittleness. This proves that synthesized pseudo-samples fail to bridge hardware gaps and inadvertently introduce domain-specific artifacts, whereas our SGC maintains a robust 78.50\% accuracy.

\noindent\textbf{2) Domain-Agnostic Robustness of Priors.} Conversely, SGC achieves robust 78.50\% (segment) and 79.94\% (subject) accuracy under extreme domain shifts. Crucially, while baselines suffer catastrophic clinical leakage (Recall $\le 61\%$), SGC maintains an exceptional 75.34\% Recall. This confirms that domain-agnostic anomaly priors immunize classifiers against hardware and demographic variations, capturing invariant pathological manifolds missed by existing paradigms.

\noindent\textbf{3) Limitations and Future Work.} Despite outperforming all zero-shot baselines, SGC's intra- vs. cross-dataset performance gap (95.19\% vs. 78.50\%) indicates that hard electrode mapping loses fine-grained spatial information. Future work will integrate Unsupervised Domain Adaptation (UDA) to seamlessly align high-dimensional manifolds across heterogeneous hardware.

\subsection{Effectiveness Beyond Augmentation: Mining Priors over Generation}
\label{subsec:beyond_aug}


To empirically validate our ``mining over generation'' philosophy, we define several internal variants: Supervised Baseline (Baseline), utilizing our pure 1D-CNN-Transformer backbone without score-guided fusion; Diffusion Augmentation (Diff-Aug), training the baseline on datasets expanded via DDPM synthesis; and Advanced Generative Heuristics, comprising Artifact Cleaning (Clean) to filter high-frequency synthesis artifacts, Hard Negative Mining (Hard Neg) to generate challenging boundary cases, and Spatial Hybridization (Hybrid) to dynamically blend spatial channels across multiple samples to enrich diversity. These exhaustive comparisons prove that complex sample synthesis fundamentally falls short of our zero-augmentation SGC.

\begin{table}[h]
  \centering
  \caption{Comparison of generative augmentation vs. explicit prior mining across datasets.}
  \label{tab:beyond_aug}
  
  \small 
  \setlength{\tabcolsep}{1.2pt} 
  \renewcommand{\arraystretch}{1.2}
  
  \resizebox{\linewidth}{!}{
  \begin{tabular}{l l c c c c c}
    \toprule
    \textbf{Dataset} & \textbf{Method} & \textbf{Acc (\%)} & \textbf{Pre (\%)} & \textbf{Rec (\%)} & \textbf{F1 (\%)} & \textbf{Sub-Acc (\%)} \\
    \midrule
    \multirow{6}{*}{\textbf{Mumtaz2016}} 
    & Baseline  & 85.57 & 84.86 & 88.13 & 84.98 & 86.52 \\
    & Hard Neg  & 91.23 & 88.15 & 95.80 & 91.14 & 92.50 \\
    & Hybrid    & 92.86 & 91.59 & 96.84 & 93.57 & 95.83 \\
    & Clean     & 93.16 & 91.19 & 94.83 & 92.20 & 93.26 \\
    & Diff-Aug  & 93.49 & \textbf{93.84} & 94.04 & 93.77 & 95.00 \\
    & \textbf{SGC (Ours)}   & \textbf{95.19} & 93.17 & \textbf{97.31} & \textbf{95.10} & \textbf{97.50} \\
    \midrule
    \multirow{6}{*}{\begin{tabular}[c]{@{}l@{}}\textbf{MODMA}\\(Zero-shot)\end{tabular}} 
    & Baseline  & 61.88 & 53.71 & 62.48 & 55.69 & 62.33 \\
    & Hard Neg  & 69.90 & 65.28 & 61.98 & 59.85 & 67.67 \\ 
    & Hybrid    & 61.81 & 60.52 & 48.94 & 51.14 & 62.00 \\
    & Clean     & 64.57 & 65.38 & 65.95 & 63.48 & 63.33 \\
    & Diff-Aug  & 66.68 & 63.02 & 57.97 & 57.22 & 67.67 \\
    & \textbf{SGC (Ours)}   & \textbf{78.50} & \textbf{70.82} & \textbf{75.34} & \textbf{69.96} & \textbf{79.94} \\
    \bottomrule
  \end{tabular}
  } 
\end{table}

\noindent\textbf{1) Performance Leap without Synthetic Noise.} As shown in Table~\ref{tab:beyond_aug}, while Diff-Aug improves upon the Baseline, it remains fundamentally inferior to SGC. Notably, SGC achieves a remarkable 95.19\% on Mumtaz2016 and maintains a robust 78.50\% in zero-shot transfer on MODMA, significantly outperforming Diff-Aug's 93.49\% and 66.68\%. Furthermore, despite exploring advanced generative heuristics, including hard negative mining (Hard Neg), hybrid spatial augmentations (Hybrid), and artifact-filtered synthesis (Clean). These variants yield only marginal intra-dataset improvements over the Baseline, failing to surpass standard Diff-Aug. More critically, none could overcome the catastrophic degradation in cross-dataset zero-shot transfer. This reinforces that explicitly extracting anomaly scores effectively bypasses the synthetic noise inherently introduced by generative data expansion.

\begin{figure}[h]
  \centering
  \begin{subfigure}{0.48\linewidth}
    \centering
    \includegraphics[width=\linewidth]{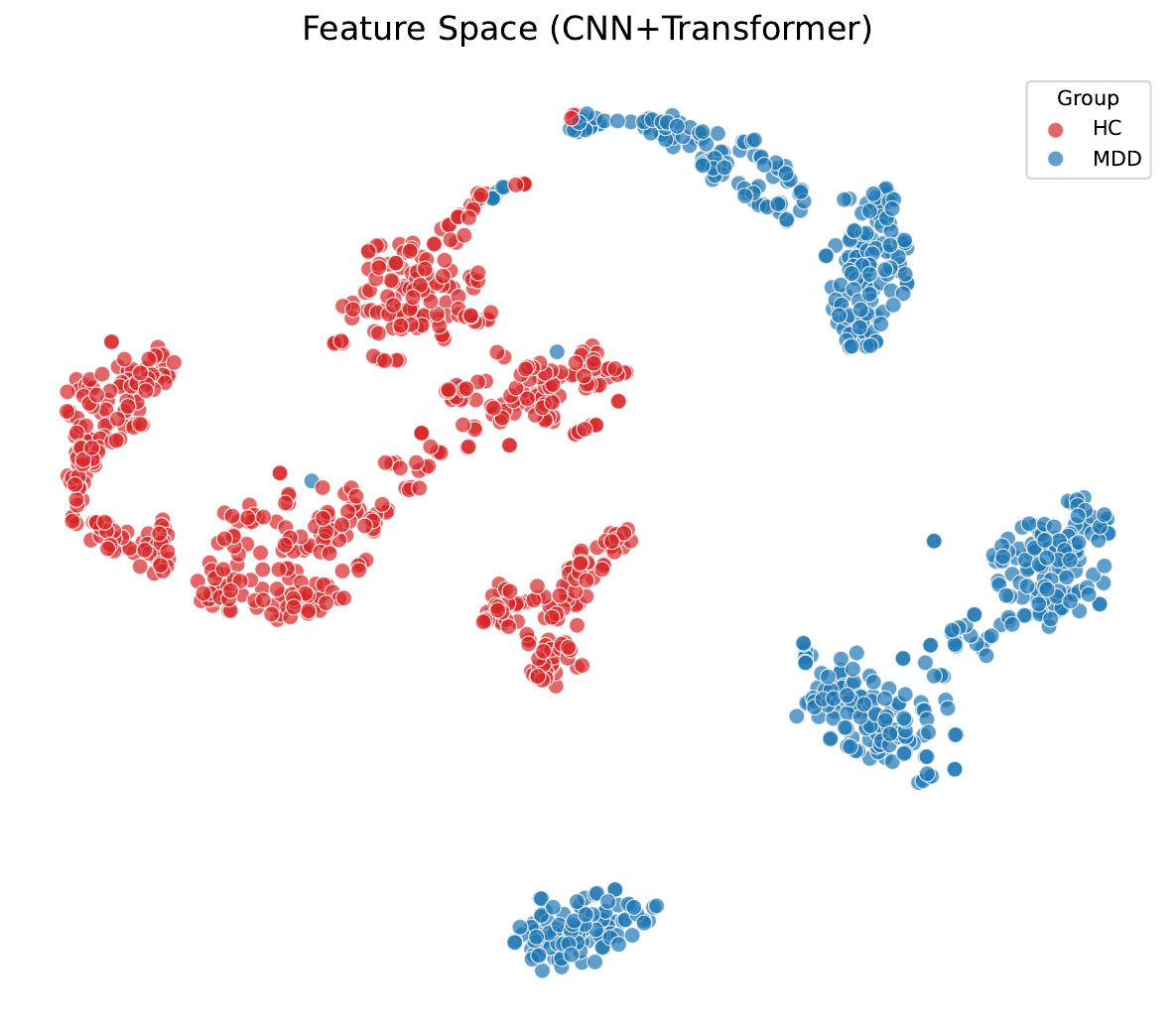}
    \caption{Baseline model}
  \end{subfigure}
  \begin{subfigure}{0.48\linewidth}
    \centering
    \includegraphics[width=\linewidth]{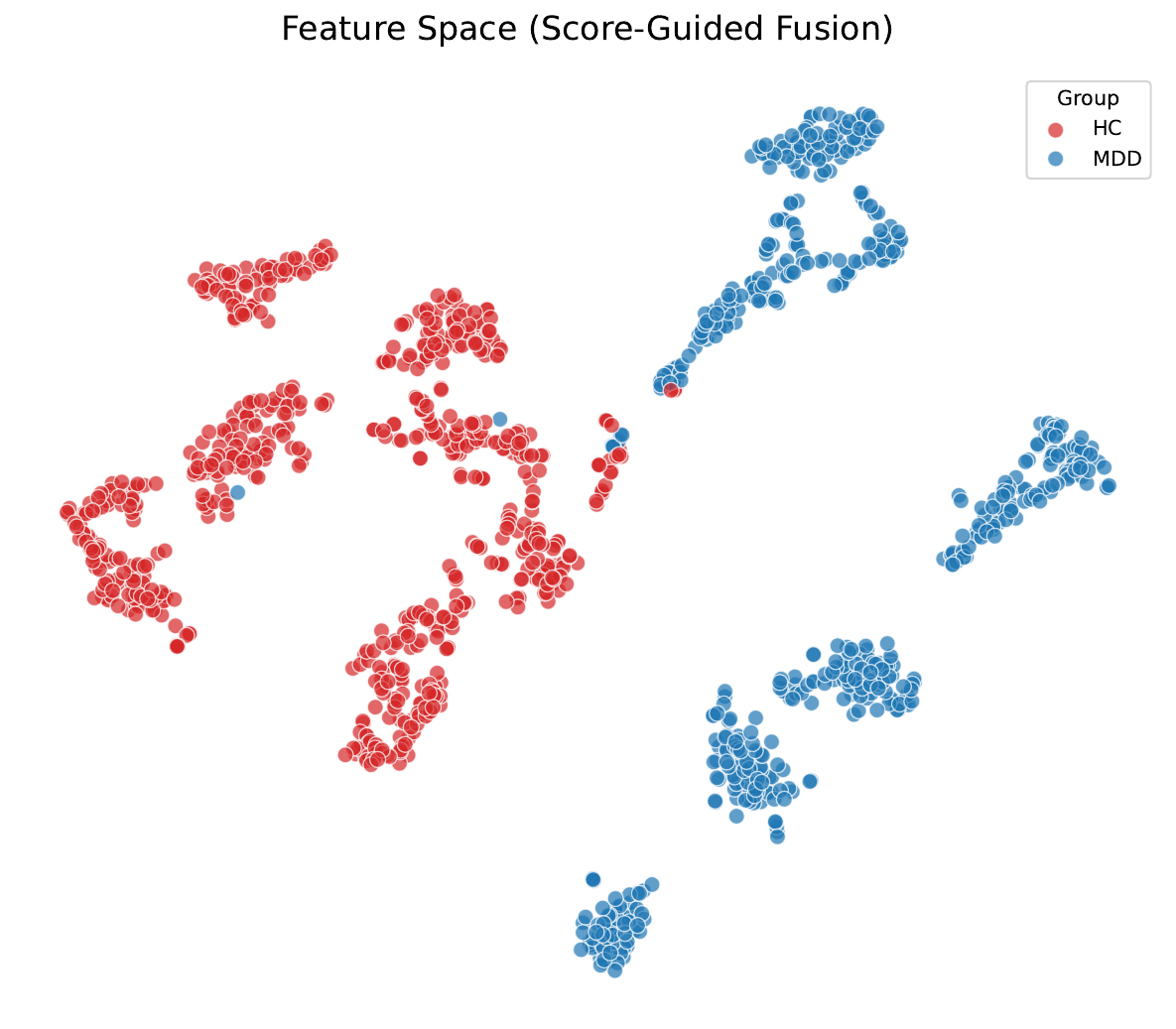}
    \caption{SGC}
  \end{subfigure}
  \caption{t-SNE visualization. SGC (b) achieves explicit feature calibration, characterized by tightened intra-class cohesion compared to the entangled Baseline (a).}
  \label{fig:tsne_comparison}
\end{figure}

\noindent\textbf{2) Reshaping the Feature Space.} Beyond numerical gains, the extracted prior serves as a potent feature calibrator. In the t-SNE projections (Fig.~\ref{fig:tsne_comparison}), the Baseline model (Fig.~\ref{fig:tsne_comparison}a) struggles with inherent EEG noise, yielding entangled boundaries. By fusing the pathological prior, SGC (Fig.~\ref{fig:tsne_comparison}b) forces the high-dimensional manifold to collapse into highly cohesive intra-class clusters while maximizing the inter-class margin.

\begin{figure}[t] 
  \centering
  \begin{subfigure}[b]{0.32\linewidth}
    \centering
    \includegraphics[width=\linewidth]{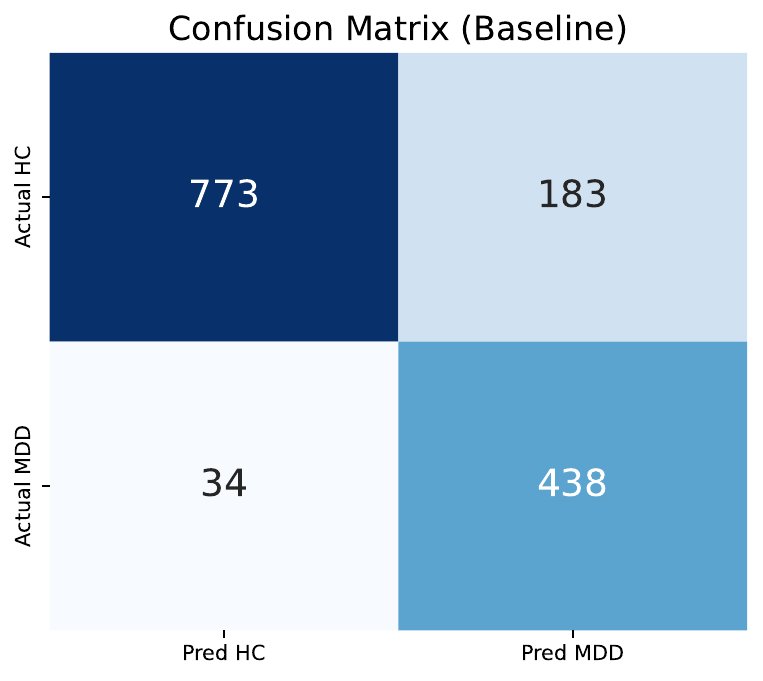}
    \caption{Baseline}
  \end{subfigure}
  \hfill 
  \begin{subfigure}[b]{0.32\linewidth}
    \centering
    \includegraphics[width=\linewidth]{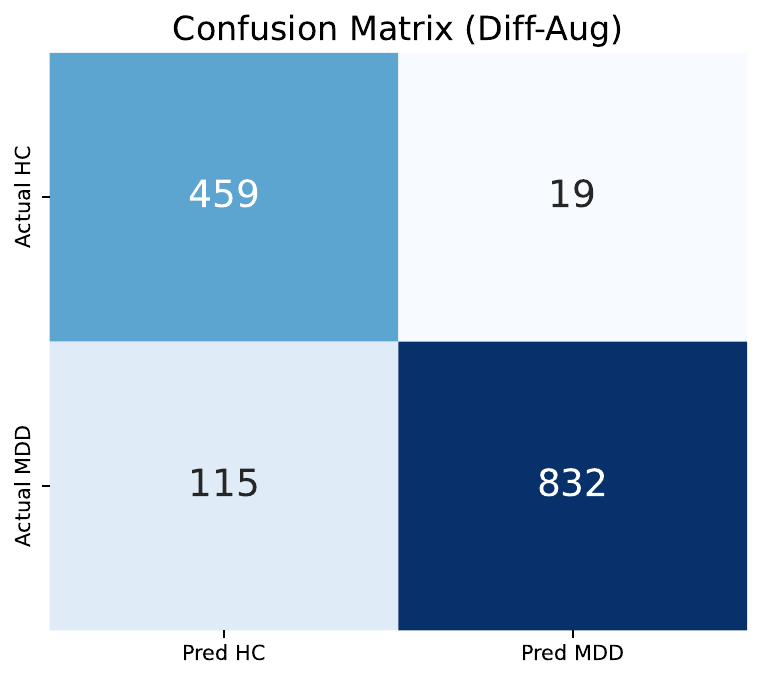}
    \caption{Diff-Aug}
  \end{subfigure}
  \hfill
  \begin{subfigure}[b]{0.32\linewidth}
    \centering
    \includegraphics[width=\linewidth]{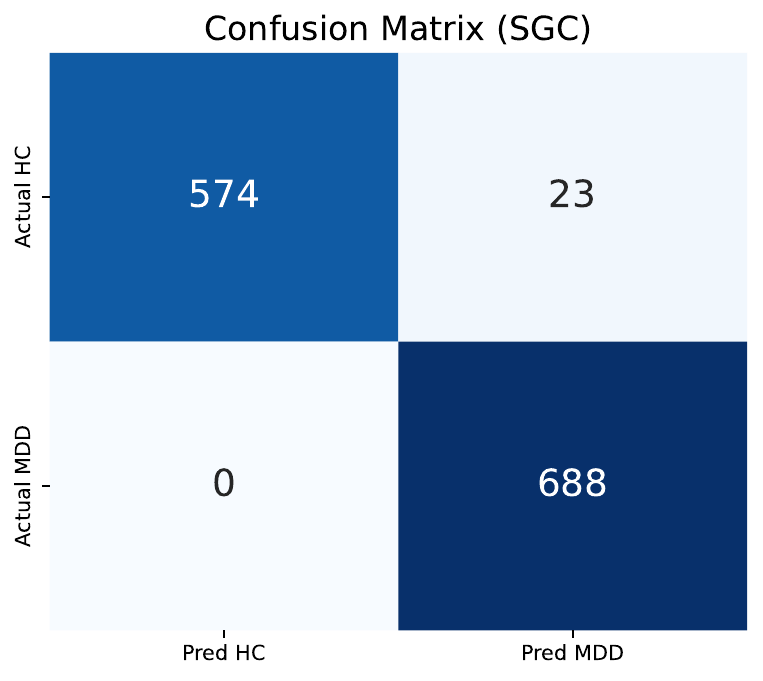}
    \caption{\textbf{SGC}}
  \end{subfigure}
  \caption{Confusion matrices on a representative fold. SGC (c) completely eliminates the clinical leakage (False Negatives) suffered by generative augmentation (b).}
  \label{fig:confusion_matrix_comparison}
\end{figure}

\noindent\textbf{3) Clinical Utility via Maximized Recall.} In psychiatric screening, misclassifying an MDD patient as healthy incurs severe clinical repercussions. Fig.~\ref{fig:confusion_matrix_comparison} contrasts the confusion matrices from a representative testing fold. While Diff-Aug (b) fails to resolve ambiguous boundary cases (missing 115 MDD patients), SGC (c) achieves a perfect clinical safety net with zero missed diagnoses. This underscores the profound translational value of SGC as a robust diagnostic fail-safe.

\subsection{Ablation Study: Complementarity of Generative Priors}
\label{subsec:ablation}

To validate the contributions of SGC's core modules, we evaluate three ablation variants across source (Mumtaz2016) and zero-shot (MODMA) domains. These include: a variant without distributional modeling (referred to as w/o Dist.), relying solely on VQ-VAE structural error; a variant without structural reconstruction (referred to as w/o Struct.), depending entirely on DDPM distribution error; and a decision fusion variant (referred to as Dec. Fusion), replacing early feature concatenation with terminal logit summation. Results are presented in Table~\ref{tab:ablation_components}.

\begin{table}[h]
  \centering
  \caption{Ablation study of SGC core components across datasets.}
  \label{tab:ablation_components}
  
  \small 
  \setlength{\tabcolsep}{1.5pt} 
  \renewcommand{\arraystretch}{1.2}
  
  \begin{tabular}{@{}llccccc@{}}
    \toprule
    \textbf{Dataset} & \textbf{Variant} & \textbf{Acc (\%)} & \textbf{Pre (\%)} & \textbf{Rec (\%)} & \textbf{F1 (\%)} & \textbf{Sub-Acc (\%)} \\
    \midrule
    \multirow{5}{*}{\textbf{Mumtaz2016}}
    & Baseline    & 85.57 & 84.86 & 88.13 & 84.98 & 86.52 \\
    & w/o Dist.   & 90.40 & 90.69 & 92.72 & 90.91 & 91.67 \\
    & w/o Struct. & 93.11 & 92.38 & 94.34 & 92.95 & 95.00 \\
    & Dec. Fusion & 93.12 & \textbf{94.10} & 91.62 & 92.00 & 94.17 \\
    & \textbf{SGC (Ours)}& \textbf{95.19} & 93.17 & \textbf{97.31} & \textbf{95.10} & \textbf{97.50} \\
    \midrule
    \multirow{5}{*}{\begin{tabular}[c]{@{}l@{}}\textbf{MODMA}\\(Zero-shot)\end{tabular}}
    & Baseline    & 61.88 & 53.71 & 62.48 & 55.69 & 62.33 \\
    & w/o Dist.   & 65.32 & 61.63 & 66.09 & 59.52 & 62.33 \\
    & w/o Struct. & 63.40 & 60.40 & 59.92 & 56.63 & 63.67 \\
    & Dec. Fusion & 69.69 & 65.45 & 68.06 & 65.30 & 69.67 \\
    & \textbf{SGC (Ours)}& \textbf{78.50} & \textbf{70.82} & \textbf{75.34} & \textbf{69.96} & \textbf{79.94} \\
    \bottomrule
  \end{tabular}
\end{table}
\noindent\textbf{1) Dual-Stream Complementarity.} Ablating either stream yields suboptimal accuracy on Mumtaz2016, whereas their integration achieves 95.19\% (Table~\ref{tab:ablation_components}). This synergy becomes critical in MODMA zero-shot transfer: individual models struggle under severe domain shifts (65.32\% and 63.40\%), yet fusion reaches 78.50\%. Notably, while distribution modeling dominates intra-dataset performance, structural quantization exhibits superior resilience against cross-domain hardware variations. This confirms that explicit morphological and implicit distributional priors capture distinct, complementary facets of pathological EEG variance.

\noindent\textbf{2) Necessity of Feature-Level Fusion.} Replacing feature concatenation with late Decision Fusion drops accuracy to 93.12\% on Mumtaz2016 and triggers a catastrophic degradation to 69.69\% on zero-shot MODMA (an 8.81\% drop). This performance collapse proves that injecting anomaly scores as high-order features enables the global attention mechanism to dynamically calibrate spatiotemporal representations against pathological priors, optimizing non-linear decision boundaries far more effectively than shallow terminal logit fusion.

\section{Conclusion}
Existing deep learning methods for EEG-based depression detection rely heavily on generative data augmentation to overcome data scarcity. However, this prevailing ``data-quantity-first'' convention inevitably introduces synthetic noise and blurs classification boundaries, compromising clinical reliability. To address this, we propose Score-Guided Classification (SGC), a novel framework that shifts the paradigm from sample generation to unsupervised prior mining. In SGC, we synergize structural anomaly metrics from a VQ-VAE and probabilistic distributional shifts from a DDPM, which constructs a potent pathological prior. This prior explicitly guides the classifier's decision boundary entirely within a zero-augmentation setting. Experiments demonstrate that SGC achieves state-of-the-art performance on the Mumtaz2016 benchmark and exhibits exceptional domain-agnostic robustness in severe zero-shot cross-dataset evaluations. By prioritizing feature integrity over synthetic quantity, SGC establishes a reliable diagnostic pathway for complex medical time-series. Future work will integrate Unsupervised Domain Adaptation (UDA) to further bridge multi-center hardware heterogeneities.

\clearpage 
\bibliographystyle{ACM-Reference-Format}
\bibliography{ref/sample-base} 

\end{document}